\documentclass[sigconf]{acmart}

\usepackage{caption}
\usepackage{subcaption}
\usepackage{bbding}
\usepackage{utfsym}
\usepackage{multirow}
\usepackage{graphicx}
\usepackage[graphicx]{realboxes}
\usepackage{amsmath}
\usepackage{amsfonts}
\usepackage{enumitem}
\usepackage{balance}
\usepackage{hyperref}

\usepackage{multirow}
\usepackage{newfloat}
\usepackage{listings}
\AtBeginDocument{%
  }

\copyrightyear{2025}
\acmYear{2025}
\setcopyright{acmlicensed}
\acmConference[MM '25] {Proceedings of the 33rd ACM International Conference on Multimedia}{October 27--31, 2025}{Dublin, Ireland.}
\acmBooktitle{Proceedings of the 33rd ACM International Conference on Multimedia (MM '25), October 27--31, 2025, Dublin, Ireland}

\begin{document}
\title{Degradation-Consistent Learning via Bidirectional Diffusion for Low-Light Image Enhancement}

\author{Jinhong He}
\email{hejh@stu.cqut.edu.cn}
\affiliation{%
  \institution{Chongqing University of Technology}
  \city{Chongqing}
  \country{China}
}

\author{Minglong Xue*}
\email{xueml@cqut.edu.cn}
\affiliation{%
  \institution{Chongqing University of Technology}
  \city{Chongqing}
  \country{China}
}


\author{Zhipu Liu}
\email{zpliu@cqut.edu.cn}
\affiliation{%
  \institution{Chongqing University of Technology}
  \city{Chongqing}
  \country{China}
}

\author{Mingliang Zhou}
\email{mingliangzhou@cqu.edu.cn}
\affiliation{%
  \institution{Chongqing University }
  \city{Chongqing}
  \country{China}
}

\author{Aoxiang Ning}
\email{ningax@stu.cqut.edu.cn}
\affiliation{%
  \institution{Chongqing University of Technology}
  \city{Chongqing}
  \country{China}
}

\author{Palaiahnakote Shivakumara}
\email{S.Palaiahnakote@salford.ac.uk}
\affiliation{%
  \institution{University of Salford}
  \city{Manchester}
  \country{UK}
}

\thanks{*Corresponding author.}

\renewcommand{\shortauthors}{Jinhong He et al.}

\begin{abstract}
Low-light image enhancement aims to improve the visibility of degraded images to better align with human visual perception. While diffusion-based methods have shown promising performance due to their strong generative capabilities. However, their unidirectional modelling of degradation often struggles to capture the complexity of real-world degradation patterns, leading to structural inconsistencies and pixel misalignments. To address these challenges, we propose a bidirectional diffusion optimization mechanism that jointly models the degradation processes of both low-light and normal-light images, enabling more precise degradation parameter matching and enhancing generation quality. Specifically, we perform bidirectional diffusion—from low-to-normal light and from normal-to-low light during training and introduce an adaptive feature interaction block (AFI) to refine feature representation. By leveraging the complementarity between these two paths, our approach imposes an implicit symmetry constraint on illumination attenuation and noise distribution, facilitating consistent degradation learning and improving the model’s ability to perceive illumination and detail degradation. Additionally, we design a reflection-aware correction module (RACM) to guide color restoration post-denoising and suppress overexposed regions, ensuring content consistency and generating high-quality images that align with human visual perception. Extensive experiments on multiple benchmark datasets demonstrate that our method outperforms state-of-the-art methods in both quantitative and qualitative evaluations while generalizing effectively to diverse degradation scenarios. \href{https://github.com/hejh8/BidDiff}{\textcolor{red}{Code}}
\end{abstract}

\ccsdesc[500]{Computing methodologies~Computational photography; Image processing}

\keywords{Low-light image enhancement, Diffusion models, Retinex prior}

\maketitle

\section{Introduction}

\begin{figure}[ht!]
        \centering
        \includegraphics[height=0.62\linewidth,width=\linewidth]{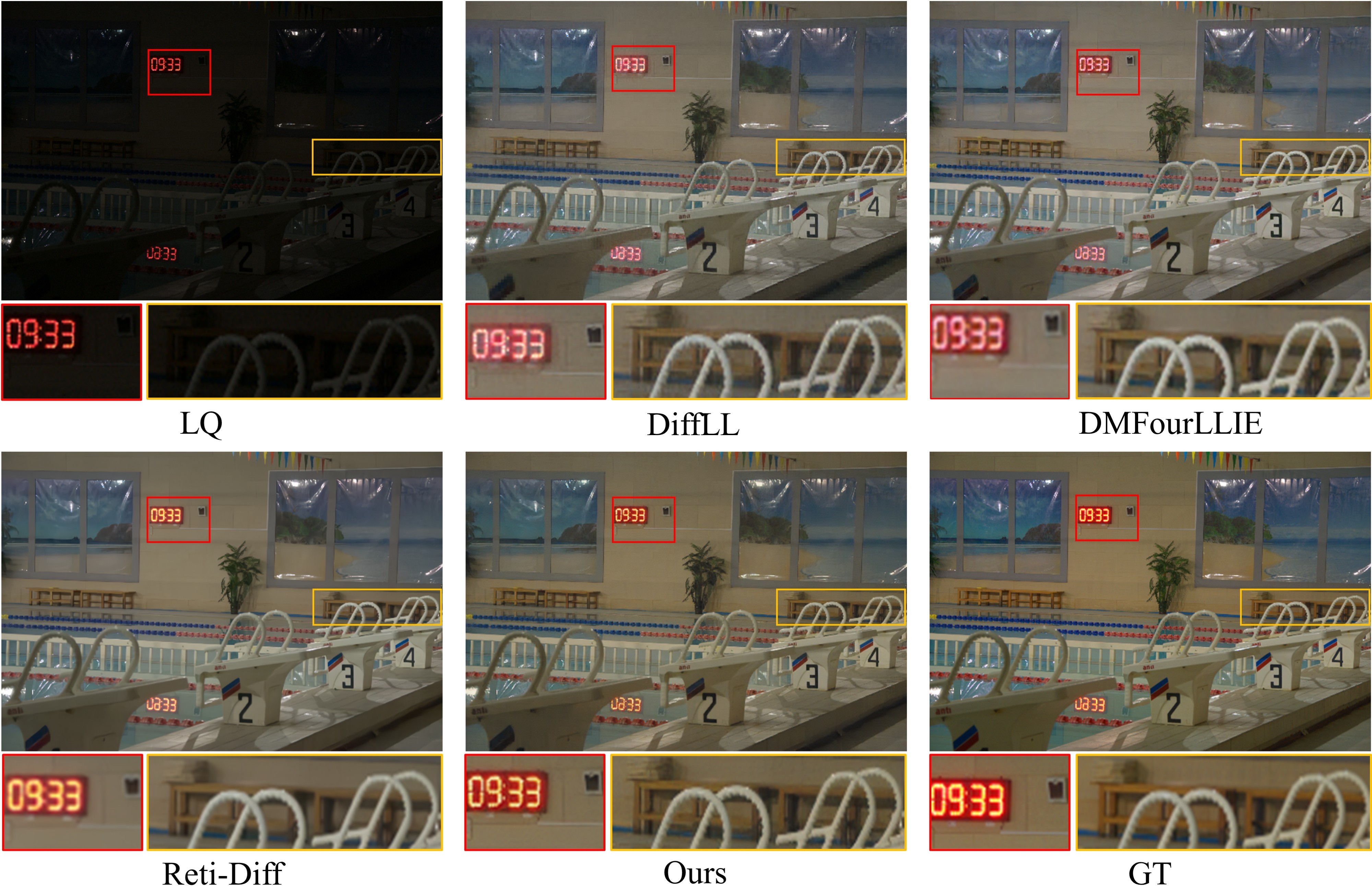}  
        \caption{Comparison of our method with competing methods on LLIE task. Our method achieves better color fidelity and generates realistic textures. Best viewed by zooming in.}
        \label{fig:1}
 \end{figure}

\begin{figure*}[t!]
        \centering
        \includegraphics[height=0.235\linewidth,width=\linewidth]{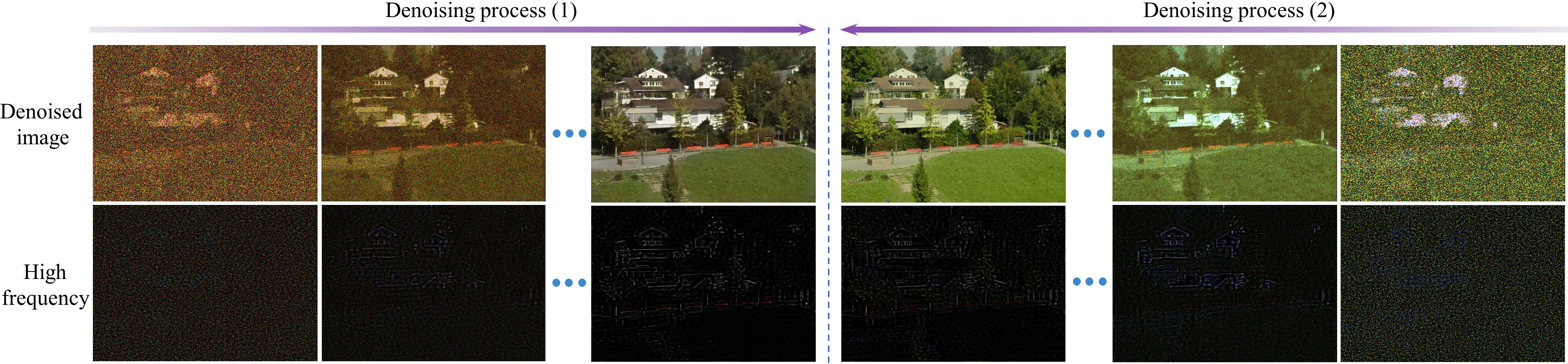}  
        \caption{Partial visualization of the denoising process. Denoising process (1) represents the results obtained through unidirectional diffusion mapping, while Denoising process (2) shows the optimized results achieved through bidirectional diffusion.}
        \label{fig:2}
\end{figure*}

The degradation of image quality in low-light environments poses a significant challenge, hindering the advancement of computer vision applications. Reduced visibility, noise interference, and color distortion severely impact critical tasks such as semantic segmentation \cite{liu2024learning}, object detection \cite{xu2021exploring}, and text detection \cite{xue2020arbitrarily}. To address this issue, low-light image enhancement (LLIE) techniques aim to restore visually informative images with natural brightness, sharp details, and accurate colors by modelling the complex transformation from degraded conditions to normal illumination. Early approaches predominantly relied on handcrafted priors, such as histogram equalization \cite{pisano1998contrast} and Retinex theory \cite{land1971lightness}, which proved effective in certain scenarios. However, these methods heavily depend on manually designed features, limiting their generalization capability. As a result, they struggle to handle complex degradation patterns, often leading to brightness imbalance, artifact amplification, and detail loss.

In recent years, deep learning has shown remarkable advantages in multi-objective collaborative optimization, including brightness restoration \cite{yang2023implicit,cai2023retinexformer}, noise suppression \cite{hou2024global}, and color correction \cite{xu2022snr,jin2023dnf}. By leveraging end-to-end feature learning and powerful nonlinear representations, these methods have opened new avenues for robust visual perception. However, despite the progress made by existing approaches \cite{zhang2024dmfourllie,liang2023iterative,wang2023fourllie} in enhancing images through improved scene rendering, they often struggle to recover fine details and accurately restore original colors, as illustrated in Fig. \ref{fig:1} (DMFourLLIE). To overcome these limitations, generative models such as Generative Adversarial Networks (GANs) \cite{jiang2021enlightengan,yang2023implicit} and Variational Autoencoders (VAEs) \cite{he2024strategic} have been explored to establish high-quality mappings between low-light and normal-light domains. These models have significantly improved perceptual quality, further driving research into generative paradigms for LLIE.

The rise of diffusion models \cite{ho2020denoising} has significantly advanced low-level vision tasks \cite{he2024zero,fei2023generative,luo2023image,ozdenizci2023restoring,xue2024low}, thanks to their powerful generative capabilities and stable performance. However, many existing methods \cite{jiang2023low,he2023reti} rely on a unidirectional mapping from low-light to normal-light domains for image-level distribution modeling, which often introduces degradation bias, leading to color distortion and structural mismatches, as seen in Fig. \ref{fig:1}. Additionally, the limitations of unidirectional mapping hinder the effective integration of constraints and prior knowledge, requiring meticulous training adjustments. This can cause the models to become overly specialized for certain tasks or datasets, diminishing their generalization across diverse degradation scenarios. Furthermore, due to the stochastic nature of the denoising process in diffusion models, the quality of constraint conditions directly influences the final image quality. This raises an intriguing question: since image degradation is essentially the inverse process of enhancement, is there an inherent correlation between the two that can be leveraged to guide the enhancement process? This consideration motivates us to explore the joint training of two diffusion processes. As illustrated in Fig. \ref{fig:2}, we visualize part of the denoising process, showing that by optimizing with both diffusion processes, the enhanced images exhibit finer content details and superior overall quality.

\begin{figure*}[t]
        \centering
        \includegraphics[height=0.32\linewidth,width=\linewidth]{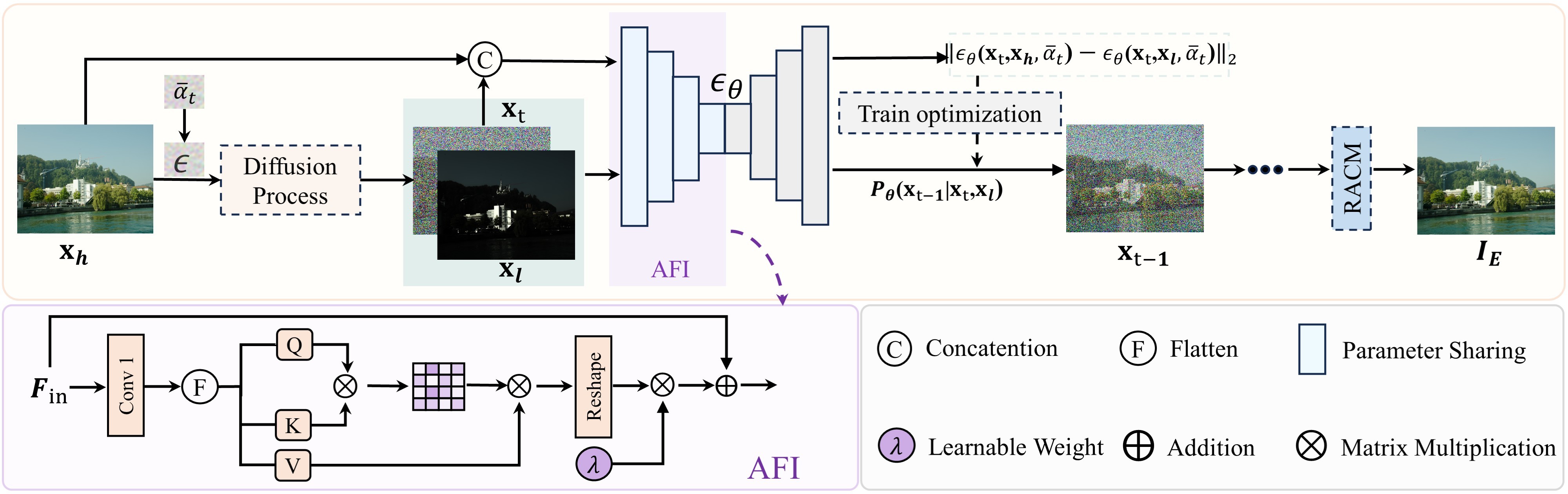}  
        \caption{Overall Pipeline of Our Proposed Method. We first generate the noisy latent variable \( x_t \) from the input normal-light image \( X_h \) and a predefined noise level \( \bar{\alpha_t} \). The normal-light image \( X_h \) and the low-light image \( X_l \) serve as conditional constraints for the H2L (Normal-to-Low light) and L2H (Low-to-Normal light) paths, respectively. These images are concatenated with \( x_t \) and fed into a U-Net for noise prediction. To enhance feature representation, we incorporate an Adaptive Feature Interaction (AFI) module. During training, we minimize the discrepancy between H2L and L2H noise estimations to optimize the L2H noise estimation network, enabling it to learn a high-quality degradation mapping. Finally, the optimized network performs reverse denoising, transforming the Gaussian noise \( x_t \) into an initial output. This output is further refined using the Reflection-Aware Correction Module (RACM) for color and illumination correction, ultimately producing the enhanced image \( I_E \).}
        \label{frame}
    \end{figure*}

With this inspiration, we propose a bidirectional diffusion optimization mechanism to facilitate consistent degradation domain learning between low-light and normal-light domains. By capturing shared degradation factors, this mechanism enhances both denoising and restoration. Specifically, during training, we perform bidirectional diffusion: from low-light to normal-light (L2H) and from normal-light to low-light (H2L). To further improve feature representation, we introduce an adaptive feature interaction block (AFI) that leverages an attention matrix to refine feature fusion. The complementary degradation understanding from L2H and H2L pathways imposes implicit symmetrical constraints on illumination attenuation and noise distribution, leading to more stable and consistent degradation parameter learning. This improves the model’s ability to perceive illumination degradation and detail loss, resulting in more robust restoration. Additionally, we propose the reflection-aware correction module (RACM) to enhance color fidelity. By extracting reflection priors through Retinex decomposition and constructing a color-aware field based on the inherent properties of the reflection map, RACM enables more accurate color restoration and effectively localizes and suppresses overexposed regions. As shown in Fig. \ref{fig:1}, our method significantly improves both global and local contrast, mitigates overcorrection in large regions, and effectively suppresses artifacts and noise amplification. Extensive experiments demonstrate the generalizability of our framework across diverse scenarios and validate its superiority over existing methods.

Overall, the main contributions of this paper can be summarized as follows:
\begin{itemize}[noitemsep,topsep=0pt]
\item We propose a bidirectional diffusion optimization mechanism that jointly models forward and reverse diffusion between low-light and normal-light domains. This approach leverages their complementarity to enforce symmetrical constraints on illumination attenuation and noise distribution, enhancing degradation perception and detail preservation. Additionally, an adaptive feature interaction block further strengthens feature representation.
\item We further develop a reflection-aware correction module (RACM), which leverages the reflection map prior derived from Retinex decomposition to guide color restoration after denoising. Additionally, it suppresses overexposed regions, ensuring content consistency and producing high-quality images that better align with human visual perception.
\item Extensive experiments demonstrate that the proposed framework exhibits strong generalization across various scenarios and outperforms existing LLIE methods.
\end{itemize}

\section{Related Work} \label{Related Works}
\subsection{Low-Light Image Enhancement}

Low-light image enhancement (LLIE) aims to reverse various degradation factors and restore clear, natural images. Existing methods can be broadly classified into traditional non-learning-based techniques \cite{fu2016weighted,guo2016lime} and deep learning-based approaches \cite{wang2024multimodal,xu2025upt,li2023embedding,zhang2024dmfourllie}. Traditional methods include histogram equalization \cite{ooi2010quadrants}, gamma correction \cite{rahman2016adaptive}, and enhancement methods based on Retinex theory \cite{ng2011total}. However, these methods rely on manually designed priors, limiting their ability to model complex lighting conditions, making them difficult to generalize and prone to amplifying noise. In recent years, deep learning methods have made significant progress in LLIE tasks, benefiting from large-scale datasets \cite{wei2018deep,yang2021sparse,wang2021seeing}. For example, RetinexNet \cite{wei2018deep} applies Retinex theory for image decomposition but suffers from some color shifts; SNRNet \cite{xu2022snr} employs a signal-to-noise ratio-aware transformer to improve color stability; CIDNet \cite{yan2025hvi} further introduces the HVI color space to separate luminance and chromaticity and enhances color fidelity via a bidirectional attention mechanism. Additionally, FourLLIE \cite{wang2023fourllie} incorporates Fourier transforms to improve robustness, while DMFourLLIE \cite{zhang2024dmfourllie} adopts a dual-stage, multi-branch Fourier design to simultaneously enhance the brightness, structure, and texture details of low-light images. Despite good performance in specific tasks, these methods are still limited in adapting to complex real-world degradation conditions, highlighting the need for more robust LLIE solutions.

\subsection{Diffusion Model for Image Restoration}

In recent years, the growing demand for high-quality image restoration has sparked increasing interest in generative paradigms. Diffusion models \cite{ho2020denoising,song2020denoising}, which generate high-quality images by optimizing the variational bound to better match the target distribution, have been widely applied to image restoration tasks \cite{xue2025unified,lin2024diffbir,yue2024efficient}. For instance, GDP \cite{fei2023generative} leverages diffusion priors to generate realistic outputs, while IR-SDE \cite{luo2023image} constructs stochastic differential equations to reverse image degradation, achieving remarkable restoration performance. WeatherDiff \cite{ozdenizci2023restoring} utilizes a local patch-based diffusion model to remove weather-related degradations. In the field of image enhancement \cite{ning2025kan,10740586,wu2024jores}, GSAD \cite{hou2024global} introduces global regularization into the diffusion process to enhance low-light images, while AnlightenDiff \cite{10740586} proposes a dynamically adjusted diffusion anchoring mechanism to ensure the authenticity of the enhancement results. Recent studies have further explored diffusion methods incorporating physical priors. For example, LightenDiffusion \cite{jiang2025lightendiffusion} integrates Retinex theory into the latent space of the diffusion model to achieve unsupervised low-light enhancement. Similarly, Reti-Diff \cite{he2023reti} extracts reflection and illumination priors within the latent space to guide image feature enhancement, thereby improving the enhancement quality. However, existing methods typically employ unidirectional degradation modeling, which may overlook certain degradation parameters, leading to local structural mismatches and pixel misalignment, ultimately affecting the final restoration quality. For a comprehensive review of this field, we refer readers to \cite{he2025diffusion}.

\section{Method} \label{Method}

The conditional diffusion model primarily consists of two steps: the forward diffusion process $q$ and the reverse sampling process $p_\theta$. The forward diffusion process is designed to generate a sequence of noisy latent variables $x_1, x_2, \dots, x_T$ through a Markov chain with a predefined variance schedule $\beta_t$. Since the noise added in the forward process is independent and follows a normal distribution with a mean of $\sqrt{1 - \beta_t} \, x_{t-1}$ and a variance of $\beta_t I$, we define $\alpha_t = 1 - \beta_t$ and $\bar{\alpha}_t = \prod_{i=1}^{t} \alpha_i$. This reformulation simplifies the forward diffusion process as follows:
\[
q(x_t | x_0) = \mathcal{N}(x_t; \sqrt{\bar{\alpha}_t} x_0, (1-\bar{\alpha}_t) I) ,
\tag{1}
\]
where \( t \in \{1, \dots, T\} \). In the reverse diffusion process, we iteratively predict the clean image \( x_0 \) by imposing the constraint of the conditional image \( y \). The reverse diffusion process is expressed as:

\[
p_\theta(x_{t-1} | x_t, y) = \mathcal{N}(x_{t-1}; \mu_\theta(x_t, y,\bar{\alpha}_t), \sigma_t^2 I) .
\label{e2}
\tag{2}
\]

In this section, we propose a bidirectional diffusion optimization strategy designed to simultaneously model the degradation processes of both low-light and normal-light images. This approach enables more precise degradation parameter matching, thereby enhancing generation quality. The overall framework is illustrated in Fig.~\ref{frame}. In the following, we provide a detailed introduction to the components of the proposed method.

\subsection{Bidirectional Diffusion Optimization}
During training, unidirectional bridging of degradation parameters may lead to the loss of certain details. To address this issue, we jointly execute the generation process of both low-light and normal-light images, enabling a more comprehensive modeling of the latent relationship between image degradation and restoration. Specifically, in the training phase, we first apply forward diffusion to the input real image \( x_h \) to obtain a purely noisy image. Then, we use the normal-light image \( x_h \) and the low-light image \( x_l \) as conditional constraints for the denoising process of the H2L and L2H paths, respectively. These images are concatenated with the noisy image before being fed into a U-Net for noise prediction. Similar to most diffusion-based methods, the UNet serves as the backbone network \( \epsilon_{\theta} \) for noise estimation. However, unlike existing approaches, we share parameters between the two bidirectional diffusion denoising paths in the encoder part of the U-Net. This design facilitates the learning of common features and enhances feature reuse capabilities. Additionally, we introduce an AFI module at specific resolution layers to further enhance feature representation. During the upsampling phase, we employ independent decoders to learn the unique degradation noise of each path, ensuring the independence of the L2H and H2L paths.

Correspondingly, in the reverse process, we follow Eq.~\ref{e2} and leverage the editing and data synthesis capabilities of the diffusion model to iteratively optimize the L2H and H2L denoising paths, thereby obtaining high-quality noise-free outputs. This process can be expressed as follows:


\[
I_E,I_D = p_\theta(x_{t-1} | x_t, y) , \ y \in \{x_l,x_h\} ,
\tag{3}
\]
where \( I_E \) represents the denoising output of the L2H diffusion path, and \( I_D \) represents the denoising output of the H2L diffusion path. During the training phase, the objective of the diffusion model is to optimize the network parameters \( \theta \) to ensure that the noise estimates for the two paths—from low-light to normal-light (L2H: \( \epsilon_{\theta}(x_t, x_l, t) \)) and from normal-light to low-light (H2L: \( \epsilon_{\theta}(x_t, x_h, t) \)) approximate standard Gaussian noise, thereby generating high-quality enhancement outputs. Furthermore, our core objective is to optimize the noise estimation network along the L2H path to improve the enhancement effect. To reinforce the learning of degradation consistency, we implicitly impose symmetric constraints on illumination attenuation and noise distribution by minimizing the differences between the noise vectors estimated by L2H and H2L, thereby achieving high-quality mappings. Based on this, we define the objective function for optimizing the diffusion model as follows:
\begin{equation}
\mathcal{L}_{diff}= \parallel \epsilon_t-\epsilon_{\theta}(x_t, x_l, t)\parallel_2 + \parallel \bar{\epsilon}_t-\epsilon_{min}\parallel_2 ,
\tag{4}
\end{equation}
where \( \bar{\epsilon}_t \) represents the noise difference \( \epsilon \) used at step \( t \) for both paths. $\epsilon_{min}=\epsilon_{\theta}(x_t, x_h, t)-\epsilon_{\theta}(x_t, x_l, t)$ represents the difference in predicted noise.

\textbf{Adaptive Feature Interaction:} As shown in Fig.~\ref{frame}, we further introduce the adaptive feature interaction (AFI) block at specific resolution layers of UNet to enhance feature representation capabilities through an attention matrix. The core idea of this module is to establish dynamic relationships between different features, thereby strengthening the feature information of the L2H and H2L paths. Given the input features \( F_{\text{in}} \), we first apply a \( 1 \times 1 \) convolutional layer to project the features, reducing computational complexity while enhancing expressiveness. Then, we flatten the projected features to generate queries (\( Q \)), keys (\( K \)), and values (\( V \)). Based on the generated \( Q \) and \( K \), we compute the attention weights via matrix multiplication and perform a weighted sum of \( V \) to obtain the enhanced output feature \( F_{\text{output}} \). To adaptively adjust the contribution of the bidirectional paths, we introduce a learnable weight \( \lambda\), allowing the model to dynamically adjust based on the input. This process can be formulated as follows:
\[
Q, K, V = Flatten(conv1(F_{in})) ,
\tag{5}
\]
\[
F_{\text{output}} = \lambda \cdot softmax(QK^T) \cdot V + F_{in} .
\tag{6}
\]



\begin{table*}[t!]
\caption{Quantitative comparison of our method with other SOTA methods in LOL-v1 \cite{wei2018deep}, LOL-v2-Real \cite{yang2021sparse} and LOL-v2-Syn \cite{yang2021sparse} datasets. ↑ (↓) denotes that, larger (smaller) values suggest better quality. The best and second performance are marked in {\textcolor[HTML]{FF0000}{red}} and {\color[HTML]{0808D4}{blue}}, respectively. Note that in the evaluation of all methods, GT Mean was not used for overall brightness adjustment.}
\renewcommand\arraystretch{0.9}
\scalebox{1.0268}{
\begin{tabular}{l|c|ccc|ccc|ccc}
\hline
                          &                             & \multicolumn{3}{c|}{LOL-v1}                                                                  & \multicolumn{3}{c|}{LOL-v2-Real}                                                            & \multicolumn{3}{c}{LOL-v2-Syn}                                                              \\ \cline{3-11} 
\multirow{-2}{*}{Methods} & \multirow{-2}{*}{Reference} & PSNR ↑                          & SSIM ↑                         & LPIPS ↓                        & PSNR ↑                          & SSIM ↑                         & LPIPS ↓                        & PSNR ↑                          & SSIM ↑                         & LPIPS ↓                        \\ \hline
DRBN \cite{yang2020fidelity}                   & CVPR'20                     & 19.774                        & 0.827                        & 0.317                                           & 19.730                        & 0.819                        & 0.252                                             & 21.627                        & 0.825                       & 0.174                 \\
MIRNet \cite{zamir2020learning}                   & ECCV'20                     & 22.122                        & 0.830                        & 0.250                        & 20.021                        & 0.820                        & 0.233                        & 22.520                        & 0.899                        & 0.110                        \\
Zero-DCE++ \cite{li2021learning}               & TPAMI'21                    & 14.682                        & 0.472                        & 0.407                        & 17.461                        & 0.490                        & 0.427                        & 17.712                        & 0.815                        & 0.168                        \\
EnlightenGAN \cite{jiang2021enlightengan}                    & TIP'21                      & 17.483                        & 0.651                        & 0.390                        & 18.641                        & 0.675                        & 0.379                        & 18.177                        & 0.793                        & 0.212                        \\
Kind++ \cite{zhang2021beyond}                   & IJCV'21                     & 21.300                        & 0.823                        & 0.207                        & 20.590                        & 0.829                        & 0.266                        & 21.170                        & 0.881                        & 0.267                        \\
SCI \cite{ma2022toward}                      & CVPR'22                     & 14.784                        & 0.526                        & 0.392                        & 17.304                        & 0.540                        & 0.345                        & 18.577                        & 0.765                        & 0.243                        \\
SNRnet \cite{xu2022snr}                   & CVPR'22                     & 24.309                        & 0.841                        & 0.262                        & 21.480                        & 0.849                        & 0.237                        & 24.138                        & 0.927                        & 0.085                        \\
UHDFour \cite{li2023embedding}                   & ICLR'23                     & 23.095                        & 0.822                        & 0.259                        & 21.785                        & 0.854                        & 0.291                        & 23.640                        & 0.900                        & 0.097                        \\
PairLIE \cite{fu2023learning}                  & CVPR'23                     & 19.735                        & 0.776                        & 0.357                        & 20.357                        & 0.782                        & 0.317                        & 19.074                        & 0.794                        & 0.230                        \\
SMG-LLIE \cite{Xu_2023_CVPR}                      & CVPR'23                     & 24.306                        & 0.813                        & 0.281                        & 22.625                        & 0.824                       & 0.287                        & 25.622                        & 0.870                        & 0.194                       \\
GDP \cite{fei2023generative}                      & CVPR'23                     & 15.904                        & 0.540                        & 0.431                        & 14.290                        & 0.493                        & 0.435                        & 16.127                        & 0.593                        & 0.279                        \\
PyDiff \cite{zhou2023pyramid}                   & IJCAI’23                    & 23.298                        & 0.858                        & 0.215                        & 23.364                        & 0.834                        & 0.208                        & 25.126                        & 0.917                        & 0.098                        \\
CLIP-LIT \cite{liang2023iterative}                 & ICCV'23                     & 12.394                        & 0.493                        & 0.397                        & 15.262                        & 0.601                        & 0.398                        & 16.190                        & 0.772                        & 0.200                        \\
Retinexformer \cite{cai2023retinexformer}               & ICCV'23       & 25.153              & 0.845               & 0.246                                              & 22.797                        & 0.840                        & 0.334      
    & 25.668                        & 0.930                        & 0.073                                             \\
NeRco \cite{yang2023implicit}                    & ICCV'23                     & 22.946                        & 0.785                        & 0.311                        & 19.234                        & 0.671                        & 0.338                        & 19.628                        & 0.799                        & 0.263                        \\
Diff-Retinex \cite{yi2023diff}                    & ICCV'23                     & 22.690                        & 0.853                        & 0.191                       & 21.190                        & 0.833                        & 0.271                      & 24.336                        & 0.921                        & 0.061                        \\
FourLLIE \cite{wang2023fourllie}                 & ACM MM'23                   & 24.150                        & 0.840                        & 0.241                        & 22.340                        & 0.841                        & 0.233                        & 24.650                        & 0.919                        & 0.099                        \\
DiffLL \cite{jiang2023low}                   & TOG'23                      & {\color[HTML]{0808D4} 26.316} & 0.844                        & 0.219                        & 22.428                        & 0.817                        & 0.191                        & 25.456                        & 0.896                        & 0.102                        \\
GSAD \cite{hou2024global}                     & NeurIPS'23                  & 23.224                        & 0.855                        & 0.193                        & 20.177                        & 0.847                        & 0.205                        & 24.098                        & 0.927                        & 0.073                        \\
AnlightenDiff \cite{10740586}            & TIP'24                      & 21.762                        & 0.811                        & 0.252                        & 20.423                        & 0.825                        & 0.267                        & 20.581                        & 0.874                        & 0.146                        \\
FourierDiff \cite{lv2024fourier}              & CVPR'24                     & 17.560                        & 0.607                        & 0.359                        & 18.673                        & 0.602                        & 0.362                        & 13.699                        & 0.631                        & 0.252                        \\
LightenDiffusion \cite{jiang2025lightendiffusion}         & ECCV'24                     & 20.188                        & 0.814                        & 0.316                        & 21.097                        & 0.847                        & 0.305                        & 21.542                        & 0.866                        & 0.188                        \\
DMFourLLIE \cite{zhang2024dmfourllie}               & ACM MM'24                   & 24.754                        & 0.859                        & 0.195                        & 22.974                        & 0.859                        & 0.187                        & 25.937                       & 0.933                       & 0.081                        \\
URetienxNet++ \cite{wu2025interpretable}           & TPAMI'25                    & 23.826                        & 0.839                        & 0.231                        & 21.967                        & 0.836                        & 0.203                        & 24.603                        & 0.927                        & 0.102                        \\
Reti-Diff \cite{he2023reti}                & ICLR'25                     & 25.132                        & {\color[HTML]{0808D4} 0.866} & 0.199                        & 22.987                        & 0.859                        & {\color[HTML]{0808D4} 0.186} & {\color[HTML]{0808D4} 27.530} & {\color[HTML]{0808D4} 0.951} & {\color[HTML]{0808D4} 0.053} \\
CIDNet \cite{yan2025hvi}                   & CVPR'25                     & 23.809                        & 0.857                        & {\color[HTML]{0808D4} 0.188} & {\color[HTML]{FF0000} 24.111} & {\color[HTML]{0808D4} 0.868} & 0.202                        & 25.129                        & 0.939                        & 0.070                        \\ \hline
Ours                      & -                           & {\color[HTML]{FF0000} 26.948} & {\color[HTML]{FF0000} 0.869} & {\color[HTML]{FF0000} 0.184} & {\color[HTML]{0808D4} 23.371} & {\color[HTML]{FF0000} 0.876} & {\color[HTML]{FF0000} 0.181} & {\color[HTML]{FF0000} 28.676} & {\color[HTML]{FF0000} 0.953} & {\color[HTML]{FF0000} 0.049} \\ \hline
\end{tabular}
}
\label{t1}
\end{table*}

\begin{figure}[t]
        \centering
        \includegraphics[height=0.3\linewidth,width=\linewidth]{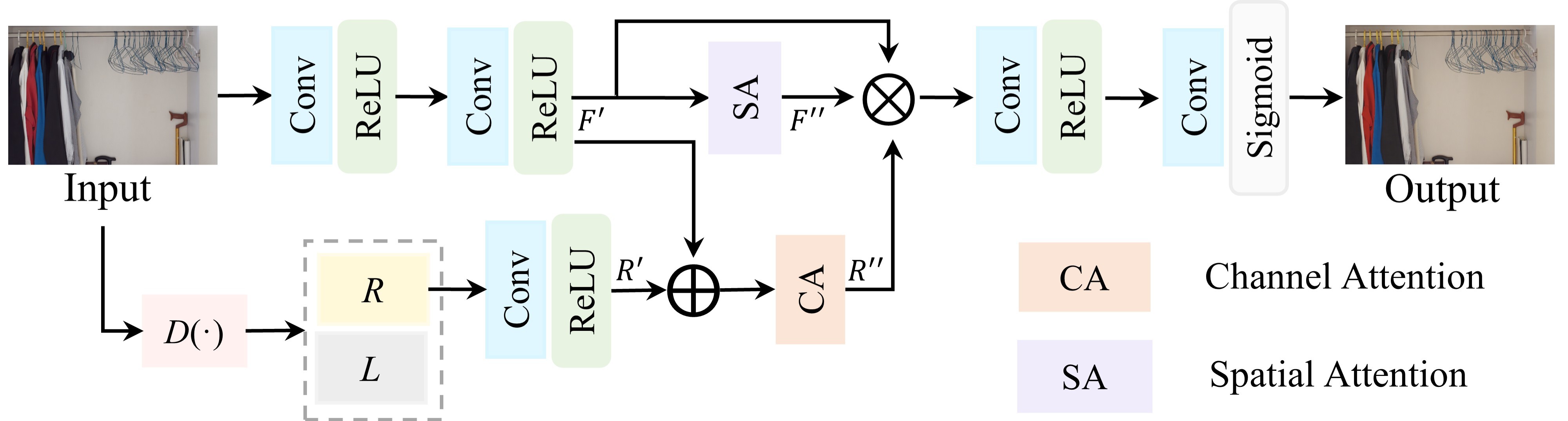}  
        \caption{The detailed architecture of our proposed RACM.}
        \label{racm}
    \end{figure}

\begin{table}[]
\caption{Quantitative Comparison with Diffusion-Based Methods on the UHD-LL \cite{li2023embedding} Dataset.}
\renewcommand\arraystretch{1.068}
\scalebox{0.99}{
\begin{tabular}{lcccc}
\hline
                          & \multicolumn{4}{c}{UHD-LL}                                                                              \\ \cmidrule(r){2-5}
\multirow{-2}{*}{Methods} & PSNR↑                         & SSIM↑                        & LPIPS↓                        & Param (M) \\ \hline
PyDiff \cite{zhou2023pyramid}                   & 26.643                        & {\color[HTML]{0808D4}{0.907}}                        & 0.153                        & 97.9      \\
DiffLL \cite{jiang2023low}                   &  25.843                       &  0.863                      &  0.237                      &  22.1         \\
Diff-Retinex \cite{yi2023diff}                   &26.590                       & 0.897                        & 0.155                        &  56.9        \\
GSAD \cite{hou2024global}                   & 26.771                     & 0.901                         & 0.178                       & 17.4      \\
AnlightenDiff \cite{10740586}            & 24.972                        & 0.881                       & 0.186                        & 37.8          \\
FourierDiff \cite{lv2024fourier}              & 18.084                       & 0.781                       & 0.225                        & 547.5      \\
LightenDiffusion \cite{jiang2025lightendiffusion}         & 23.854                        & 0.864                        & 0.187                        & 27.8      \\
Reti-Diff \cite{he2023reti}                &  {\color[HTML]{0808D4}{26.917}}                    & 0.905                       &  {\color[HTML]{0808D4}{0.138} }                      & 26.1      \\ \hline
Ours                      & {\color[HTML]{FF0000} 28.052} & {\color[HTML]{FF0000} 0.923} & {\color[HTML]{FF0000} 0.105} & 33.5          \\ \hline
\end{tabular}
}
\label{t2}
\vspace{-10pt}
\end{table}

\begin{figure*}[t]
        \centering
        \includegraphics[height=0.37\linewidth,width=\linewidth]{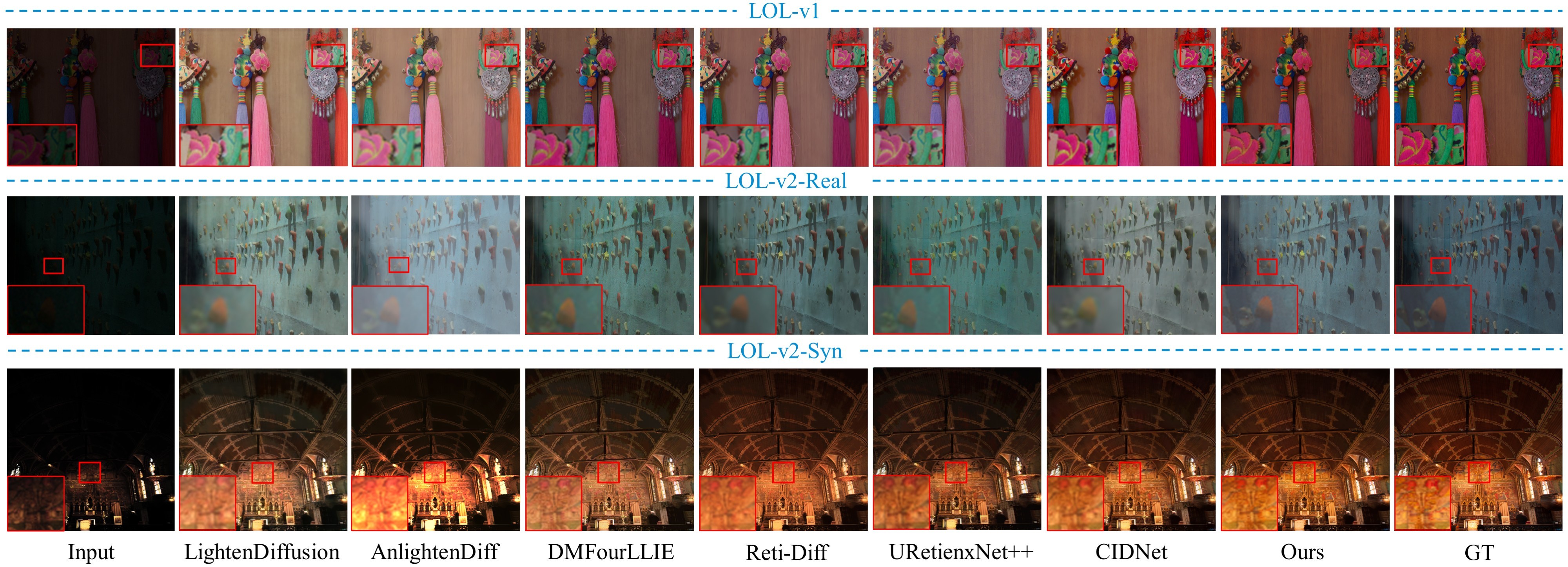}  
        \caption{Visual comparison on LOL-v1, LOL-v2-real, and LOL-v2-Synthesis datasets. Our proposed method effectively enhances visibility while maintaining superior image details and preserving the original colors. Best viewed by zooming in.}
        \label{pair1}
    \end{figure*}

\begin{table}[t]
\caption{MUSIQ scores on DICM \cite{lee2013contrast}, LIME \cite{guo2016lime}, and MEF \cite{ma2015perceptual} datasets. Larger scores suggest better quality. The best and second performance are marked in {\textcolor[HTML]{FF0000}{red}} and {\color[HTML]{0808D4}{blue}}, respectively. "AVG" denotes the average MUSIQ scores across these three datasets}
\renewcommand\arraystretch{1.068}
\scalebox{1.088}{
\begin{tabular}{lcccc}
\hline
Methods          & DICM     & MEF                           & LIME                          & AVG                           \\ \hline
SNRNet \cite{xu2022snr}          & 53.594                        & 56.920                        & 52.165                        & 54.226                        \\ \hline
CLIP-Lit \cite{liang2023iterative}        & {\color[HTML]{0808D4} 63.072} & 62.357                        & 62.865                        & 62.765                        \\
UHDFour \cite{li2023embedding}         & 59.238                        & 59.419                        & 58.125                        & 58.927                        \\
GDP \cite{fei2023generative}             & 55.608                        & 58.978                        & 57.300                        & 57.295                        \\
Diff-Retinex \cite{yi2023diff}          & 61.853                        & 61.089                        & 60.422                        & 61.121                        \\
DiffLL \cite{jiang2023low}          & 58.264                        & 62.601                        & 57.617                        & 59.494                        \\
GSAD \cite{hou2024global}           & 59.876                        & 57.649                        & 58.755                        & 58.760                        \\
FourierDiff \cite{lv2024fourier}     & 57.118                        & 57.405                        & 56.829                        & 57.117                        \\
LightenDiffusion \cite{jiang2025lightendiffusion} & 55.812                        & 58.669                        & 61.147                        & 58.543                        \\
DMFourLLIE \cite{zhang2024dmfourllie}      & 61.795                        & 61.529                        & {\color[HTML]{0808D4} 63.723} & 62.349                        \\
URetienxNet++ \cite{wu2025interpretable}   & 62.334                        & 60.330                        & 62.056                        & 61.573                        \\
Reti-Diff \cite{he2023reti}       & 62.656                        & {\color[HTML]{0808D4} 62.814} & 63.376                        & {\color[HTML]{0808D4} 62.949} \\
CIDNet \cite{yan2025hvi}          & 59.512                        & 62.794                        & 59.008                        & 60.438                        \\ \hline
Ours             & {\color[HTML]{FF0000} 63.186} & {\color[HTML]{FF0000} 63.022} & {\color[HTML]{FF0000} 64.503} & {\color[HTML]{FF0000} 63.570} \\ \hline
\end{tabular}
}
\label{t3}
\end{table}

\subsection{Reflection-Aware Correction Module}
The Retinex theory suggests that the original image can be decomposed into the reflection component \( R \) and the illumination component \( L \), and that the \( R \) remains consistent under different lighting conditions. This can be expressed as:
\[
I = R \odot L .
\tag{7}
\]

In practical applications, the process of removing illumination effects is often hindered by noise interference and errors in illumination estimation, which may cause the reflection map \( R \) to expose certain overexposed regions. Most existing methods focus on reconstructing \( R \) to preserve content details and color constancy, while optimizing the lighting component \( L \) to enhance the image. However, in complex real-world scenarios, some critical details may be hidden within the lighting component, leading to information loss and degraded image quality. Therefore, in our approach, using the reflection map \( R \) as prior information to guide the recovery of both content and color, while also suppressing exposed areas, seems to be a more suitable direction.

%

As shown in Fig. \ref{racm}, we integrate reflectance priors to construct the Reflection-Aware Correction Module (RACM), which adaptively corrects image brightness and color using Channel Attention (CA) and Spatial Attention (SA). Specifically, we first perform Retinex decomposition \( D(\cdot) \) on the input image to obtain the reflectance prior \( R \), which provides reliable color and structural information. Then, we apply a combination of convolution and ReLU to both the input image and the reflectance prior \( R \) to extract deep embedded features \( F^{\prime} \) and \( R^{\prime} \). Based on the extracted features, we employ channel attention to distinguish true color information from overexposure noise and adjust the color information accordingly as $R^{\prime \prime} = CA(R^{\prime}+F^{\prime})$. Meanwhile, we apply spatial attention to the input image features \( F^{\prime} \) to enhance the response to high-frequency components such as edges and textures, preventing local blurring caused by illumination adjustments as $F^{\prime \prime} = SA( F^{\prime} ) $.

Next, we combine channel attention and spatial attention weights and use Hadamard multiplication to refine the input image features (\(F^{\prime} \cdot F^{\prime \prime} \cdot R^{\prime \prime}\)), achieving two types of adaptive adjustments in the feature space: strengthening features in texture-rich regions that are not marked as overexposed while enhancing global color consistency, and reducing the feature amplitudes of highly exposed regions. Finally, after reconstructing the feature dimensions through a convolutional block, we obtain a high-quality image with further corrected and optimized brightness and color.

\subsection{Network Optimisation}
In addition to the target function \( L_{\text{diff}} \) used to optimize the noise estimation network, we further consider perceptual information and regional differences to improve image quality both qualitatively and quantitatively. Our loss function consists of four key components:
\[
\mathcal{L}_{total} = \omega_1 \mathcal{L}_{diff} + \omega_2 \mathcal{L}_{content} + \omega_3 \mathcal{L}_{structural} ,
\tag{8}
\]
where \( \omega \) represents the loss weights, which are empirically set as \( \omega_1, \omega_2, \omega_3 = [1, 0.3, 1] \). Specifically, the structural loss $\mathcal{L}_{structural} = (1 - SSIM(I_E, x_h))$ represents the structural similarity loss between the L2H path generation result \( I_E \) and the reference image $x_h$. \( \mathcal{L}_{content} \) represents the content loss between the final generated results of the two paths in bidirectional diffusion and the reference image, which aims to minimize the content difference between the generated image and the reference image:
\[
\mathcal{L}_{content} = \sum_{l=0}^{4} \left( \tau_1 \left\| \Phi_l(I_E) - \Phi_l(x_h) \right\|_2 + \tau_2 \left\| \Phi_l(I_{D}) - \Phi_l(x_l) \right\|_2 \right) ,
\tag{9}
\]
where \( \tau \) represents the weights for the content loss of the L2H and H2L paths, we set them as \( 0.9 \) and \( 0.1 \). \( \Phi_l \) denotes the feature extracted from the \( l \)-th layer of the ResNet101 model.

\section{Experiments}
\subsection{Experimental Settings}

\textbf{Implementation Details.} We train our method on two NVIDIA RTX 3090 GPUs using PyTorch with the Adam optimizer (1000 epochs, batch size 8, initial learning rate $2\times10^{-4}$ ). The H2L path is used only during training and is omitted at inference. After training, high-quality results are achieved with just 10 denoising steps. Experiments are conducted on LOL-v1 \cite{wei2018deep}, LOL-v2-Real \cite{yang2020fidelity}, LOL-v2-Synthetic \cite{yang2020fidelity}, UHD-LL \cite{li2023embedding} (paired) and DICM \cite{lee2013contrast}, LIME \cite{guo2016lime}, MEF \cite{ma2015perceptual} (unpaired) datasets, with PSNR, SSIM \cite{wang2004image}, LPIPS \cite{zhang2018unreasonable} for paired, and MUSIQ \cite{ke2021musiq} for unpaired evaluation. Additional results are included in the Appendix.

\begin{figure*}[ht!]
        \centering
        \includegraphics[height=0.352\linewidth,width=\linewidth]{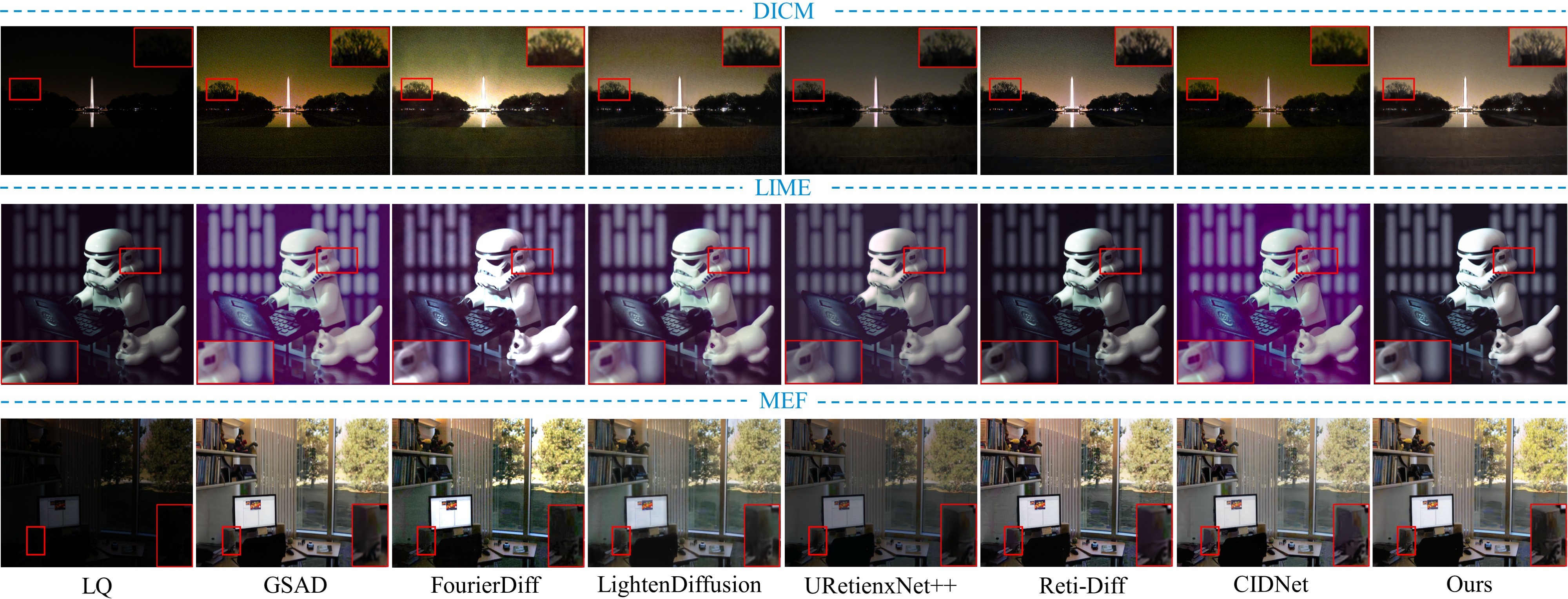}  
        \caption{Visual comparison on the unpaired datasets DICM, LIME, and MEF. Best viewed by zooming in.}
        \label{unpair}
    \end{figure*}


\subsection{Quantitative Evaluation}
We obtained the quantitative results of other methods by utilizing their official pre-trained models and running their publicly available code. As shown in Table \ref{t1}, our method consistently and significantly outperforms all comparison methods. Specifically, in terms of PSNR evaluation, our method achieves remarkable improvements of 0.632 dB and 1.146 dB on the LOL-v1 and LOL-v2-Synthesis datasets, respectively, while maintaining the second-best performance on the LOL-v2-Real dataset. For SSIM evaluation, our method achieves superior performance, ranking first across all datasets, with an outstanding SSIM score of 0.876 on the LOL-v2-Real dataset. Furthermore, for the perceptual metric LPIPS, our method also achieves the best scores across all datasets. To further validate the effectiveness of bidirectional diffusion, we compare our method with diffusion-based methods on the UHD-LL dataset. As shown in Table \ref{t2}, our method achieves the best performance across all metrics. These impressive results indicate that the optimization through bidirectional diffusion enables learning a more precise degradation domain mapping, verifying the effectiveness of our strategy in addressing the LLIE task.

As shown in Table \ref{t3}, we report the evaluation results on unpaired datasets. A higher MUSIQ score indicates better naturalness and visual quality of the image. It is evident that our method achieves the best MUSIQ scores across all datasets. This further demonstrates the superiority and generalization ability of our approach.


\subsection{Qualitative Evaluation}
Fig. \ref{pair1} presents the qualitative results on paired datasets. A detailed comparison with competing methods demonstrates that our approach exhibits superior enhancement capabilities. Specifically, previous methods suffer from poor global contrast and blurred details. For instance, in LOL-v1, Reti-Diff and URetinexNet++ exhibit color distortions, while AnlightenDiff produces over-brightened results. Although DMFourLLIE and CIDNet achieve relatively good enhancement, their background details and overall color clarity are still inferior to ours, which more closely aligns with the ground truth (GT). In LOL-v2-Real, almost all methods suffer from severe color deviations, resulting in an overall greenish contrast. In LOL-v2-Synthesis, LightenDiffusion introduces noise, leading to blurry images. Additionally, AnlightenDiff and URetinexNet++ struggle with insufficient brightness and overexposure issues.

Furthermore, Fig. \ref{unpair} presents the qualitative results on unpaired datasets. Reti-Diff and URetinexNet++ fail to produce fully brightened visual results, while GSAD and CIDNet cause severe color distortions. In contrast, our method preserves rich details and effectively mitigates color discrepancies, achieving a visual effect that closely resembles the GT. More details are provided in \textbf{Appendix}.

\begin{table}[]
\caption{Quantitative comparison of image inpainting on the CelebA-HQ dataset and denoising on the BSD68 dataset.}
\renewcommand\arraystretch{1.168}
\scalebox{0.83}{
\begin{tabular}{lcccccc}
\hline
                          & \multicolumn{3}{c}{Inpainting}                                                              & \multicolumn{3}{c}{Denoising}                                                                 \\
                          \cmidrule(r){2-4} \cmidrule(r){5-7}
\multirow{-2}{*}{Methods} & PSNR ↑                          & SSIM ↑                         & LPIPS ↓                        & PSNR ↑                          & SSIM ↑                         & LPIPS ↓                        \\ \hline
Restormer \cite{zamir2022restormer}                & {\color[HTML]{0808D4} 29.881} & {\color[HTML]{0808D4} 0.914} & 0.070                        & {\color[HTML]{0808D4} 27.246} & 0.762                        & {\color[HTML]{0808D4} 0.215} \\
NAFNet \cite{chen2022simple}                    & 29.731                        & 0.907                        & 0.085                        & 27.160                        & {\color[HTML]{0808D4} 0.768} & 0.216                        \\
IR-SDE \cite{luo2023image}                    & 27.557                        & 0.884                        & 0.061                        & 24.821                        & 0.640                        & 0.232                        \\
DA-CLIP \cite{luo2024controlling}                   & 29.277                        & 0.901                        & {\color[HTML]{0808D4} 0.042} & 24.333                        & 0.571                        & 0.269                        \\ \hline
Ours                      & {\color[HTML]{FF0000} 30.712} & {\color[HTML]{FF0000} 0.927} & {\color[HTML]{FF0000} 0.040} & {\color[HTML]{FF0000} 27.258} & {\color[HTML]{FF0000} 0.788} & {\color[HTML]{FF0000} 0.198} \\ \hline
\end{tabular}
}
\label{irs}
\end{table}

\begin{figure}[t!]
        \centering
        \includegraphics[height=0.45\linewidth,width=\linewidth]{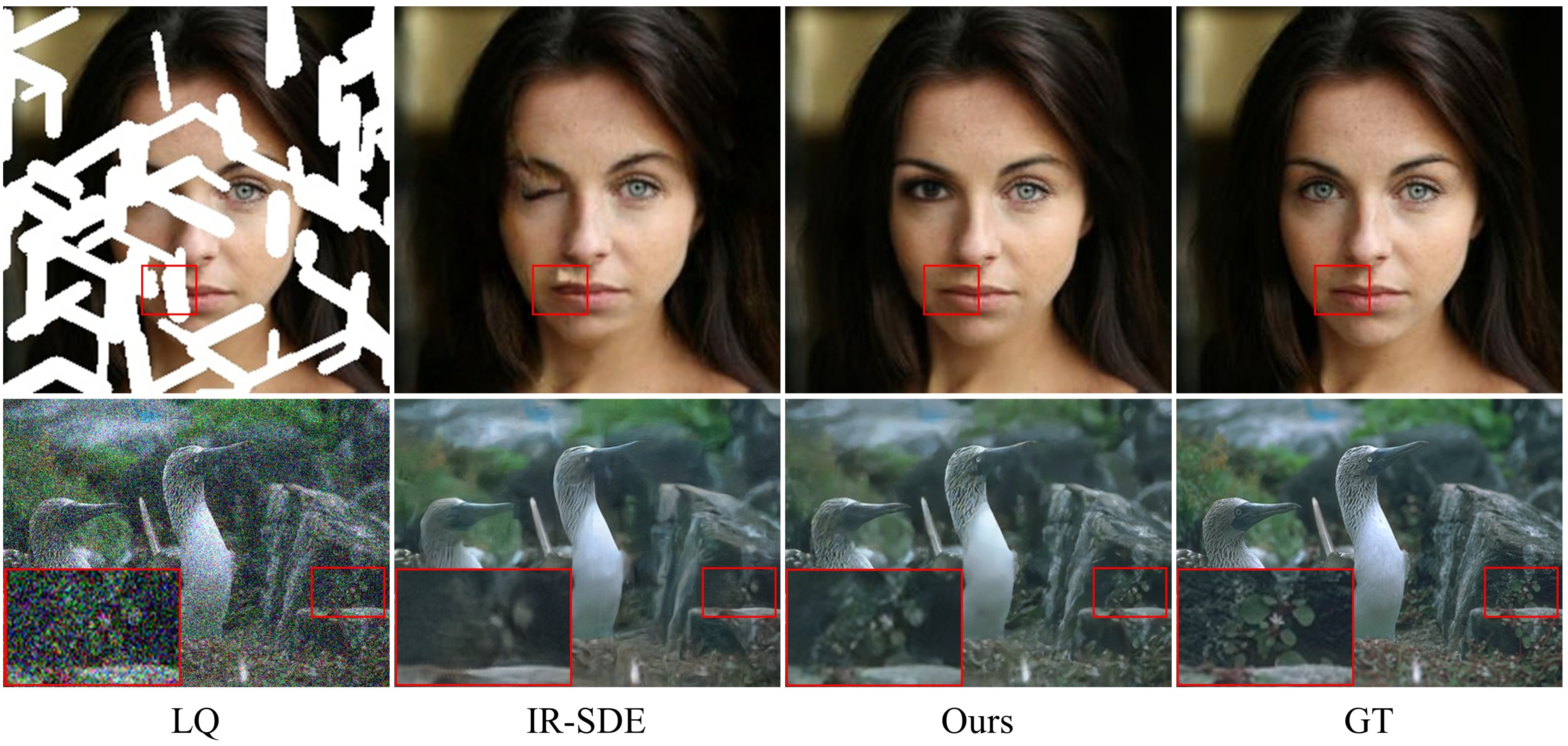}  
        \caption{Visual comparison with the diffusion-based method IR-SDE on image inpainting and image denoising tasks. Best viewed by zooming in.}
        \label{fig:ir}
    \end{figure}

\subsection{Cross-Task Validation}
To further evaluate the generalization capability of our method, we conducted cross-task validation on various low-level vision tasks. Specifically, we applied the model to tasks beyond low-light image enhancement, such as image denoising on the CBSD68 dataset and image inpainting on the CelebA-HQ dataset. CBSD68 \cite{martin2001database} consists of 68 images for denoising, where Gaussian noise with a noise level of 50 is added. CelebA-HQ \cite{huang2018introvae} is evaluated using 100 images with thin mask segmentation, following the RePaint \cite{lugmayr2022repaint} setting.

The experimental results, as shown in Table \ref{irs}, demonstrate that our approach achieves competitive performance across different tasks. This indicates that the bidirectional diffusion optimization effectively learns robust feature mappings, enabling our model to generalize well across multiple image restoration tasks. Furthermore, as illustrated in Fig. \ref{fig:ir}, we compare our method with the diffusion-based method IR-SDE. The results demonstrate that our method achieves superior visual perception quality.

\subsection{Ablation Study}
\begin{figure}[t]
        \centering
        \includegraphics[height=0.6\linewidth,width=\linewidth]{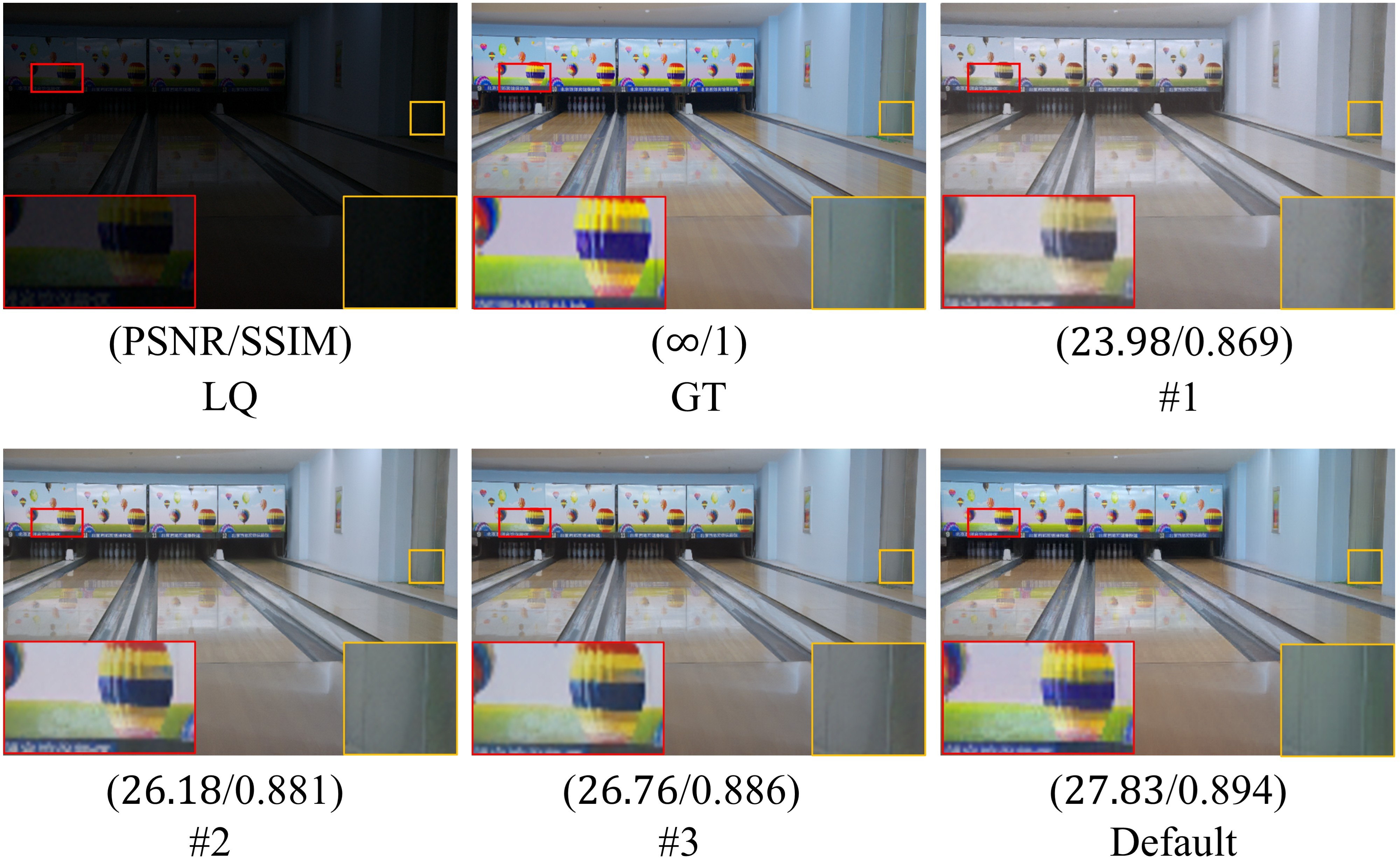}  
        \caption{Visualization of the component ablation in the proposed method. The full set performs best. Meanwhile, both PSNR and SSIM exhibit significant improvements, highlighting their importance in image enhancement.}
        \label{fig:as}
    \end{figure}

\begin{table}[t]
\renewcommand\arraystretch{1.068}
\centering
\caption{Ablation study on different model settings in the LOL-v1 dataset. 'H2L' represents the use of the normal-light to low-light path for bidirectional diffusion. 'Default' indicates the complete model setting.}
\scalebox{1.068}{
\begin{tabular}{c|ccc|ccc}
\hline
index & H2L & AFI & RACM & PSNR↑   & SSIM↑  & LPIPS↓     \\ \hline
\#1   & \usym{2613}      & \usym{2613}       & \usym{2613}      & 23.577 & 0.835 & 0.211 \\
\#2   & \usym{1F5F8}      & \usym{2613}       & \usym{2613}       & 25.648 & 0.858 & 0.193  \\
\#3   & \usym{1F5F8}      & \usym{1F5F8}      & \usym{2613}       & 26.163 & 0.861 & 0.190  \\
Default    & \usym{1F5F8}      & \usym{1F5F8}      & \usym{1F5F8}      & \textbf{26.948} & \textbf{0.869} & \textbf{0.184}  \\ \hline
\end{tabular}
}
\label{tab:33}
\end{table}

{\bf The Effectiveness of different components.} Ablation experiments on a LOL-v1 dataset, we verified their effectiveness. As shown in Table \ref{tab:33}, Among them, \#1 represents the basic setting without any modifications. \#2 introduces the normal-light to low-light (H2L) path for bidirectional diffusion. \#3 further incorporates the adaptive feature interaction (AFI) module. \#4 employs the full configuration. We observe that compared to the baseline setting \#1, \#2 achieves significant performance improvements, particularly in PSNR and SSIM evaluation, with gains of 2.071 dB and 0.23. This demonstrates the necessity of the bidirectional diffusion strategy for degradation learning. To further enhance feature representation, we adopt configuration \#3, which further optimizes model performance. Finally, by RACM, we obtain the most complete configuration and achieve the best evaluation results. Specifically, compared to \#3, it further improves the PSNR and SSIM scores by 0.785 dB and 0.08, respectively, while also enhancing perceptual measurement in LPIPS. This highlights the crucial role of the RACM component in color and illumination correction. Furthermore, the visual results in Figure \ref{fig:as} provide additional validation of the effectiveness of each component. For instance, after incorporating the bidirectional diffusion strategy (\#2), the image exhibits significant improvements in color fidelity and detail preservation. With the further refinement and correction introduced by the AFI and RACM modules, the enhanced image gradually achieves visually friendly optimization.

\noindent{\bf The Effectiveness of the Loss Function.} We validate the effectiveness of the proposed loss function by systematically removing each component and reporting the quantitative results in Table \ref{tab:loss}. As shown in the second column, removing the diffusion loss $\mathcal{L}_{diff}$ significantly degrades model performance, highlighting its crucial role in generation quality and the importance of the bidirectional diffusion strategy. The content loss $\mathcal{L}_{content}$ leads to noticeable changes in the generated images, particularly in distortion metrics, where PSNR and SSIM improve by 1.571 dB and 0.16, respectively. The structural loss $\mathcal{L}_{structural}$ is designed to further aid in reconstructing image details and stabilizing diffusion-based content generation; thus, its removal results in performance degradation.

\begin{table}[t!]
\centering
\renewcommand\arraystretch{1.0}
\caption{Ablation studies of the loss function terms.}
\scalebox{0.958}{

\begin{tabular}{lcccc}
\hline
                    & \multicolumn{4}{c}{Loss Settings}                                         \\
                    \cmidrule(r){2-5}
\multirow{-2}{*}{Metrics} & w/o $\mathcal{L}_{diff}$ & w/o $\mathcal{L}_{content}$  & w/o $\mathcal{L}_{structural }$  & Default                       \\ \hline
PSNR ↑                & 25.648       & 25.378        & 26.643        & \textbf{ 26.948} \\
SSIM ↑               & 0.851        &  0.853       & 0.857        &\textbf{ 0.869}  \\
LPIPS ↓               & 0.207             & 0.213        & 0.187        &\textbf{ 0.184}  \\ \hline
\end{tabular}
}
\label{tab:loss}
\end{table}

\section{Conclusion}
In this paper, we propose a novel bidirectional diffusion optimization mechanism for low-light image enhancement. By jointly modelling the diffusion processes between low-light and normal-light domains, our approach implicitly establishes symmetrical constraints on illumination attenuation and noise distribution, enabling more precise degradation parameter estimation and improved detail preservation. Additionally, we introduce an adaptive feature interaction block to enhance feature representation and design a reflection-aware correction module (RACM) to refine color restoration while effectively suppressing overexposure artifacts. Extensive experiments on multiple benchmark datasets demonstrate the effectiveness of our method, showing that it consistently outperforms state-of-the-art approaches in both quantitative and qualitative evaluations. Furthermore, the strong generalization ability of our framework across diverse degradation scenarios highlights its potential for real-world applications.

\begin{acks}
This work was supported in part by the National Natural Science Foundation of China under Grant 62176027; the Science and Technology Research Program of Chongqing Municipal Education Commission, China (Grant No. KJQN202401106); the Chongqing Talent under Grant, China (cstc2024ycjh-bgzxm0082); the Central University Operating Expenses under Grant, China (2024CDJGF-044).
\end{acks}

\bibliographystyle{ACM-Reference-Format}
\balance
\bibliography{sample-base}

\end{document}